\documentclass[11pt,a4paper]{article}
\usepackage[hyperref]{emnlp2018}
\usepackage{times}
\usepackage{latexsym}
\usepackage{url}
\usepackage{amsmath}
\usepackage{tabularx}
\usepackage{booktabs}
\usepackage{multicol}
\usepackage{multirow}
\usepackage[pdftex]{graphicx} 
\usepackage{color}

\aclfinalcopy 

\title{Weighted Global Normalization for Multiple Choice Reading Comprehension over Long Documents}

\author{Aditi Chaudhary$^{1*}$,\hspace{0.2em} Bhargavi Paranjape$^{1*}$,\hspace{0.2em} Michiel de Jong$^{2}$ 
\thanks{\hspace{2mm} All authors contributed equally}\\
 Carnegie Mellon University$^1$ \hspace{0.3em} University of Southern California$^2$\\
 \texttt { \{aschaudh, bvp\}@cs.cmu.edu,  \hspace{0.2em}  msdejong@usc.edu}
 }

\date{}

\begin{document}
\maketitle
\begin{abstract}

Motivated by recent evidence pointing out the fragility of high-performing span prediction models, we direct our attention to multiple choice reading comprehension. In particular, this work introduces a novel method for improving answer selection on long documents through weighted global normalization of predictions over portions of the documents. We show that applying our method to a span prediction model adapted for answer selection helps model performance on long summaries from NarrativeQA, a challenging reading comprehension dataset with an answer selection task, and we strongly improve on the task baseline performance by +36.2 Mean Reciprocal Rank.
\end{abstract}

\section{Introduction}
The past years have seen increased interest from the research community in the development of deep reading comprehension models, spurred by the release of datasets such as SQuAD. \cite{squad}. For a majority of these datasets, the top performing models employ span prediction, selecting a span of tokens from the reference document that answers the question. Such models have been very successful; the best model on the SQuAD leaderboard approaches human performance \cite{qanet}. However, this strong performance may be deceptive. \cite{adversarial} show that inserting lexically similar adversarial sentences into the passages sharply reduces performance. 

One possible reason for this disparity is that standard span prediction is an easy task. The information required to evaluate whether a span is the correct answer is often located right next to the span. \citet{kernelselection} transform the SQuAD dataset into a sentence selection task where the goal is to predict the sentence that contains the correct span. They achieve high accuracy on this task using simple heuristics that compare lexical similarity between the question and each sentence individually, without additional context.
Selecting an answer from a list of candidate answers that are lexically dissimilar to the context makes it more challenging for models to retrieve the relevant information. For that reason, we focus on reading comprehension for answer selection. 

Another common weakness of reading comprehension datasets is that they consist of short paragraphs. This property also makes it easier to locate relevant information from the context. Realistic tasks require answering questions over longer documents. 

Building on \cite{clarkgardner}, we propose a weighted global normalization method to improve the performance of reading comprehension models for answer selection on long documents. First, we adapt global normalization to the multiple-choice setting by applying a reading comprehension model in parallel over fixed length portions (chunks) of the document and normalizing the scores over all chunks. Global normalization encourages the model to produce low scores when it is not confident that the chunk it is considering contains the information to answer the question. Then we incorporate a weighting function to rescale the contribution of different chunks. In our work we use an Multilayer Perceptron over the scores and a TF-IDF heuristic as our weighting function, but more complex models are possible.  

We experiment on the answer selection task over story summaries from the recently released NarrativeQA \cite{narrativeqa} dataset. It provides an interesting and a challenging test bed for reading comprehension as the summaries are long, and the answers to questions often do not occur in the summaries. We adopt the three-way attention model \cite{triattention}, an adapted version of the BiDAF \cite{bidaf} span prediction model, in order to evaluate our method. We show that straightforward application of the answer selection model to entire summaries fails to outperform the model where the context is removed, demonstrating the weakness of current reading comprehension (RC) models . Inspired by \citet{chen2017reading},  we show that using TF-IDF to reduce context and applying global normalization on top of the reduced context, both significantly improve the performance. We observe that incorporating TF-IDF scores into the model with weighted global normalization helps improve performance more than either individually.

We view our contribution as twofold:
\begin{itemize}
\itemsep-0.3em 
\item We introduce a novel weighted global normalization method for multiple choice question answering over long context documents. 
\item We improve over the baseline of NarrativeQA answer selection task by a large margin, setting a competitive baseline for this interesting and challenging reading comprehension task.
\end{itemize}
\section{Model Architecture}
\label{model}
While span prediction models are not directly applicable to the answer selection task, the methods used to represent the context and question carry over. We base our model on the multiple choice architecture in \cite{triattention}. Taking inspiration from the popular BiDAF architecture \cite{bidaf}, the authors employ three-way attention over the context, question, and answer candidates to create and score answer representations. This section outlines our version of that architecture, denoted by \textit{T-Attn}.

\paragraph{Word Embedding Layer.} We use pre-trained word embedding vectors to represent the tokens of query, context and all candidate answers.

\paragraph{Attention Flow Layer.} The interaction between query and context is modeled by computing a similarity matrix. This matrix is used to weigh query tokens to generate a query-aware representation for each context token. We compute query-aware-answer and context-aware-answer representations in a similar fashion. The representation of a token $u$ in terms of a sequence $\mathbf{v}$ is computed as:
\begin{gather}
\mathit{Attn_{seq}}(u, \{v_i\}_{i=1}^n) = \sum_{i=1}^{n}\alpha_iv_i\\
\alpha_i = \mathit{softmax}(\mathit{f}(\mathbf{W}u)^T\mathit{f}(\mathbf{W}v_i))
\end{gather}
where $\mathit{f}$ is ReLU activation.

\paragraph{Modeling Layer.} We encode the query, context and the candidate answers by applying a Bi-GRU to a combination of the original and query/context-aware representations.  Consequently, we obtain:-
\begin{gather}
\mathbf{h_q} = \text{Bi-GRU}({w_i}_{i=1}^{|Q|})\\
\mathbf{h_c} = \text{Bi-GRU}([{w_i;w_i^q}]_{i=1}^{|C|})\\ \mathbf{h_a} = \text{Bi-GRU}([{w_i;w_i^q;w_i^c}]_{i=1}^{|a|})
\end{gather}
where $w_i$ are token embeddings, and $w_i^q$ and $w_i^c$ are query-aware and context-aware token representations respectively.
\paragraph{Output Layer.} The query  and each candidate answers is re-weighted by weights learnt through a linear projection of the respective vectors. Context tokens are re-weighted by taking a bilinear attention with \textbf{q}.
\begin{gather}
\mathbf{q}=Attn_{self}\mathbf{h_q}^{|Q|}\\ \mathbf{a} = Attn_{self}\mathbf{h_a}^{|a|} \\
\mathbf{c} = Attn_{seq}(\mathbf{q},{h_c}^{|C|})\\
\mathit{Attn_{self}}(u_i) = \sum_{i=1}^{n}\mathit{softmax}(\mathbf{W}^\top u_i)u_i
\end{gather}   where $\mathbf{h_x}$ is the hidden representation from the respective modeling layers.
We adapt the output layer used in \cite{triattention} for multiple-choice answers by employing a feed-forward network ($ffn$) to compute scores $S_a$ for each candidate answer $a$. The formulation is:-
\begin{gather}
\mathbf{l_{a^q}} = \mathbf{q}^\top \mathbf{a} \quad ; \quad
\mathbf{l_{a^c}} = \mathbf{c}^\top \mathbf{a}\\
S_a = ffn([\mathbf{l_{a^q}};\mathbf{l_{a^c}}])
\end{gather}
Standard cross-entropy loss over all answer candidates is used for training.

\section{Evaluation}
In this section, we will first discuss the different methods, including our proposed method, used for handling the challenge of longer context documents. Since existing methods don't apply to our task of answer selection as is, we also discuss the adjustments we made to these existing methods for comparing it with our proposed approach. For our task, we used the \textit{T-Attn} model, described in Section \ref{model}, as our standard reading comprehension model.
\subsection{Existing methods}

\paragraph{Baseline.} The baseline is set by using the reading comprehension model,  \textit{T-Attn} model in our case, on the entire long context document. 
\paragraph{Heuristic context reduction.} A simple method to make reading comprehension on longer contexts manageable is to use heuristic information retrieval methods, like TF-IDF as used by  \cite{chen2017reading},  to reduce the context first and then apply any of the standard reading comprehension models to this reduced context. 

 We divide the summaries (the long context) into chunks comprising of approximate 40 tokens. These chunks are then ranked by their TF-IDF score with either the question (during validation and testing) or with the question and the gold answer (during training). We then apply the reading comprehension model to the $K$ top ranked chunks. We experiment with $k=1,5$. Reducing the context in this way make it easier for the reading comprehension model to locate relevant information, but runs the risk of eliminating important context.

\paragraph{Global Normalization.} \citet{clarkgardner} improve span prediction over multiple paragraphs by applying a reading comprehension model to each paragraph separately and globally normalizing span scores over all paragraphs.

This technique does not directly apply to the case of answer selection, as the correct answer is not a span and hence not tied to a specific paragraph. We implement an adjusted version of global normalization, in which we apply a reading comprehension model to each paragraph (in our case, each chunk $j$) separately, and sum all chunk scores over the answer candidates. The probability of answer candidate $i$ being correct, given $m$ chunks is

\begin{equation*}
p_i = \frac{\sum_{j=1}^m e^{s_{ij}}}{\sum_{j=1}^m \sum_{i=1}^n e^{s_{ij}}}
\label{gn}
\end{equation*}
where $s_{ij}$ is score given by  $j^{th}$ chunk to $i^{th}$ answer candidate.

Normalizing in this manner encourages the model to produce lower scores for paragraphs that do not contain sufficient information to confidently answer the question.
\subsection{Proposed method}
\paragraph{Weighted global normalization.}

The global normalization method relies on the reading comprehension model to learn what paragraphs contain useful information, rather than using the TF-IDF heuristic alone to eliminate paragraphs, which runs the risk of mistakenly culling useful context. 

We incorporate the TF-IDF scores $h_j$ into reading comprehension model to re-weigh each chunk as follows:
\begin{equation}
p_i = \frac{ \sum_{j=1}^m z_j e^{s_{ij}}}{\sum_{j=1}^m z_j \sum_{i=1}^n e^{s_{ij}}}
\label{wgn}
\end{equation}
where $z_j=h_j$. These static scores for each chunk can be substituted with a learned function of these scores. For the purpose of demonstration, we use a simple MultiLayer Perceptron as the learning function, but it can be easily replaced with any function of choice. Hence, in Equation \ref{wgn},  $z_j = \mathbf{W_2}(ReLU(\mathbf{W_1}[h, s_{:,j}]))$, and we refer to this model as \textit{Wt-MLP}. 

\begin{figure}
\includegraphics[width=0.45\textwidth]{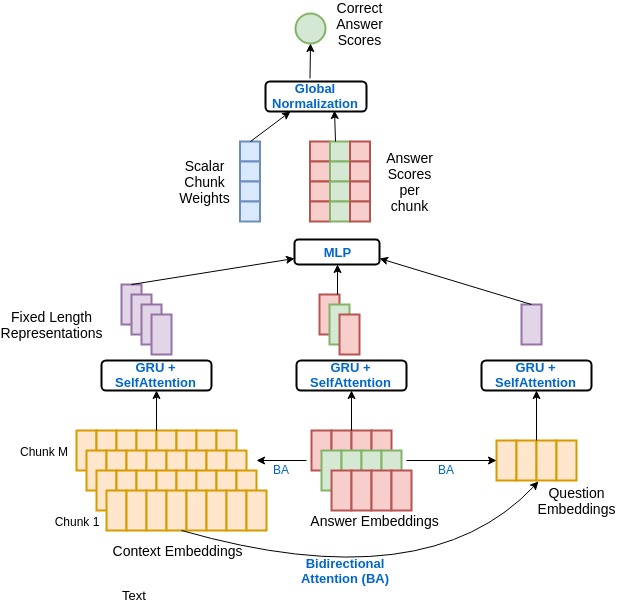}
\caption{Weighted Global Normalization for Answer Selection}
\label{architecture}
\end{figure}

The adapted Tri-Attention architecture with weighted global normalization for multiple choice answers, is shown in Figure \ref{architecture}.

\begin{table*}[ht]
\centering
\begin{tabular}{ p{6.5cm}p{3cm}p{2cm}p{2cm}}
\toprule
\textbf{Model} &\textbf{Training Context} & \textbf{Validation} & \textbf{Test}\\
&  & MRR & MRR\\
\midrule  
Weighted-Global-Norm (WGN-MLP)  & Top 5 chunks & \textbf{0.631} & \textbf{0.621}\\
Weighted-Global-Norm (WGN)  & Top 5 chunks & 0.625 & 0.613\\
\hline
Global-Norm (GN) & Top 5 chunks & 0.573 & 0.568\\
  & 5 uniform chunks & 0.545 & 0.531\\
  \hline
Vanilla Tri-Attention (T-Attn) &  Top 1 chunk & 0.601 & 0.591\\
 & Full Summary & 0.523 & 0.516\\
 & No Context & 0.525 & 0.522 \\
\hline
NarrativeQA Baseline (ASReader) &Full Summary &0.269 & 0.259 \\
\midrule
 \end{tabular}
\caption{\label{expts} Mean Reciprocal Rank by model and size of training context. The evaluation context consists of the top 5 chunks for all models except top 1 chunk, for which we evaluate on a single chunk.}
\label{results}
\end{table*}

\begin{table}[ht]
  \begin{tabular}{p{1.5cm}p{1cm}p{1cm}p{1cm}p{1cm}}
    \toprule
    \multirow{2}{*}{\textbf{Model}} &
     \multirow{2}{*}{\textbf{Train}} &
      \multicolumn{3}{c}{\textbf{Validation}} \\
        &  & \textbf{Top-1} & \textbf{Top-5}& \textbf{Full} \\
        \midrule
        WGN & Top-5 & 0.595 & 0.625 & 0.611\\
        GN & Top-5 & 0.605	& 0.573 & 0.518 \\
        T-Attn& Top-1 & 0.601 & 0.551 & 0.471 \\
       \midrule
  \end{tabular}
  \caption{Ablation: Performance of different models on different validation context sizes}
  \label{ablation}
\end{table}

\section{Experiments}

\subsection{Data}

The NarrativeQA dataset consists of 1572 movie scripts and books (\href{https://www.gutenberg.org/}{Project Gutenberg}) in the public domain, with 46765 corresponding questions. Each document is also accompanied by a corresponding summary extracted from Wikipedia. In this paper, we focus on the summaries as our dataset which have an average length of 659 tokens. For comparison, the average context size in the SQuAD dataset is less than 150 tokens. As described in \cite{narrativeqa}, candidates for a question comprise of answers \footnote{after removing duplicate answers}  to all the questions for a document. There are approximately 30 answer candidates per document. 






 
\subsection{Implementation Details}
We split each summary into chunks of approximately 40 tokens, respecting sentence boundaries. We use 300 dimensional pre-trained \cite{pennington2014glove} \textit{GloVe}  word embeddings, that are held fixed during training. We use a hidden dimension size of 128 in recurrent layers and 256 in linear layers. We use a 2-layer Bi-GRU in the modeling layer and a 3-layer feed-forward network for the output layer. The \textit{Wt-MLP} in weighted global normalization is 1 layer deep and takes following features as input:TF-IDF scores, max, min, average and standard deviation of answer scores for all chunks. A 0.2 dropout is used with Adam optimizer and a learning rate 0.002. The models converged within 10 epochs.

\section{Results and Discussion}
Table \ref{results} reports results for our main experiments. We find that \textit{T-Attn} without any context beats the NarrativeQA baseline by a large margin, suggesting the need for a stronger baseline. Surprisingly, \textit{T-Attn} on full summaries performs no better than the No Context setting, implying that the model is unable to extract relevant information from a long context.  

Providing the model with a reduced context of the top TF-IDF scored chunk (\textit{Top 1 chunk}) leads to a significant gain (by +7). Global normalization also helps; uniformly sampling 5 chunks with global normalization (GN) yields a modest improvement over no-context, and taking the top 5 TF-IDF chunks rather than randomly sampling further improves performance, though not up to the level of the reduced context.

Both global normalization and TF-IDF scoring appear to provide a useful signal on the relevance of chunks. We found that combining the two in the form of weighted global normalization (WGN-MLP) outperforms both the globally normalized (by +6) and reduced context (by +3) experiments. The global normalization helps the model better tolerate the inclusion of likely, but not certainly, irrelevant chunks, while the weighting allows it to retain the strong signal from the TF-IDF scores.

Table \ref{ablation} provides insight into the effect of global normalization. Global normalization models perform similarly to vanilla Tri-Attention trained on the top chunk when evaluated on the top chunk at test time, but degrade less in performance when evaluated on more chunks. Weighted global normalization in particular suffers only a minor penalty to performance from being evaluated on entire summaries. This effect may be even more pronounced on datasets where the top TF-IDF chunk is less reliable.  


\section{Conclusion and Future Work}
This work introduces a method for improving answer selection on long documents through weighted global normalization of predictions over chunks of the documents. We show that applying our method to a span prediction model adapted for answer selection aids performance on long summaries in NarrativeQA and we strongly improve over the task baseline. In this work, we used a learned function of candidate and TF-IDF scores as the weights, but in principle the weighting function could take any form. For future work, we intend to explore the use of neural networks that takes into account the context and query to learn weights.

\bibliography{emnlp2018}
\bibliographystyle{acl_natbib_nourl}
\end{document}